\documentclass[letterpaper]{article}

\usepackage{times}
\usepackage{url}
\usepackage[utf8]{inputenc}
\usepackage[small]{caption}
\usepackage{graphicx}
\usepackage{amsmath}
\usepackage{booktabs}
\usepackage{algorithm}
\usepackage{algorithmic}
\usepackage{wrapfig}
\usepackage{multirow}
\usepackage{tikz}
\usetikzlibrary{calc,arrows,positioning,shapes,shadows,spy,snakes,plotmarks,matrix,fit,backgrounds}
\usepackage{authblk}

\newcommand{\sanet}{{\sf SA-Net}}

\title{\sanet{}: Robust State-Action Recognition for Learning from Observations}

\author{Nihal Soans, Ehsan Asali, Yi Hong, and Prashant Doshi}
\affil{Dept. of Computer Science, University of Georgia, Athens GA 30606}
\date{}
\begin{document}

\maketitle
\thispagestyle{empty}
\pagestyle{empty}

\begin{abstract}

Learning from observation (LfO) offers a new paradigm for transferring task behavior to robots. LfO requires the robot to observe the task being performed and decompose the sensed streaming data into sequences of state-action pairs, which are then input to LfO methods. Thus, recognizing the state-action pairs correctly and quickly in sensed data is a crucial prerequisite. We present \sanet{} a deep neural network architecture that recognizes state-action pairs from RGB-D data streams. \sanet{} performs well in two replicated robotic applications of LfO -- one involving mobile ground robots and another involving a robotic manipulator -- which demonstrates that the architecture could generalize well to differing contexts. Comprehensive evaluations including deployment on a physical robot show that \sanet{} significantly improves on the accuracy of the previous methods under various conditions. %one of which utilizes traditional image processing and segmentation.  
\end{abstract}

%------------------------------------------------------------
\section{Introduction}
\label{sec:intro}
%------------------------------------------------------------

Recent robot learning methods for learning from demonstration~\cite{argall2009survey,Arora19:Survey} allow a transfer of preferences and policy from the expert performing the task to the learner. These methods have allowed the learning robots to successfully perform difficult acrobatic aerial maneuvers~\cite{abbeel2007application}, carry out nontrivial manipulation tasks~\cite{pollard2004generalizing}, penetrate patrols~\cite{bogert2014multi}, and merge autonomously into a congested freeway~\cite{Nishi19:Merging}. One way by which this transfer occurs is the learner simply observing the expert perform the task. 
Observing the expert engaged in the task is expected to yield trajectories of state-action pairs, which is then input to the algorithms that drive some of these methods. Consequently, recognizing the expert's state and action accurately from observations is crucial for the learner. If the learner is a robot, its observations are sensor streams. Very likely, these will be streams from camera and range sensors yielding RGB and depth (RGB-D) data. Thus, the learning robot must recognize sequences of state-action pairs quickly and accurately from RGB-D streams. This is a key component of the learning from observation (LfO) pipeline.

\begin{figure}[!t]
\centerline{\includegraphics[width=3.5in]{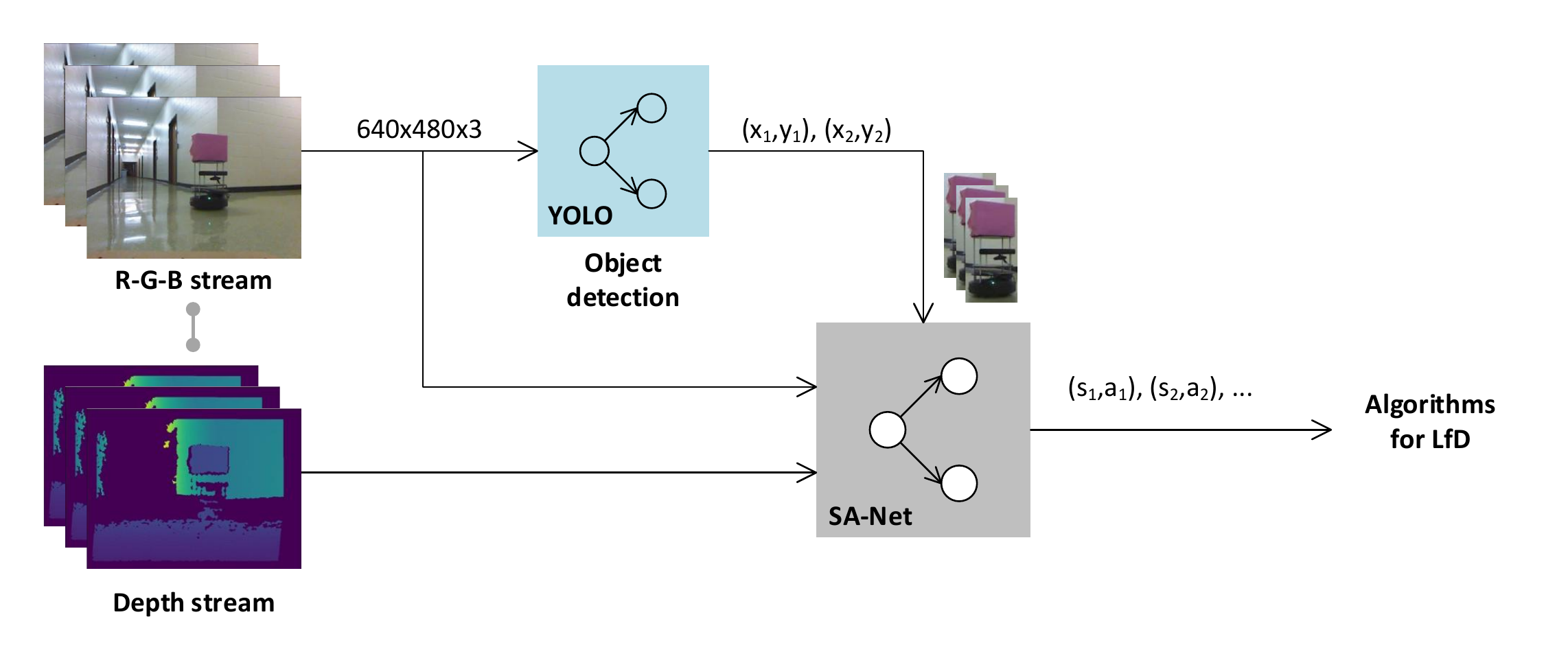}}
\caption{Overview of the I/O of \sanet{} for state-action recognition from RGB-D streams of a TurtleBot patrolling a corridor.}
\label{fig:overview}
\end{figure}

In this paper, we present \sanet{}, a deep neural network that recognizes state-action pairs from RGB-D data streams with a high accuracy. This supervised learning method offers a general deep learning alternative to the current adhoc techniques, which often rely on problem-specific implementations using OpenCV. Figure~\ref{fig:overview} gives an overview of how \sanet{} is deployed. \sanet{} aims to recognize from a sensor stream, the expert's state and action. The state is often the 2D or 3D coordinates in a global reference frame and the orientation. 
%For example, the state of a ground mobile robot is its 2D coordinates and the angle it is facing as measured counterclockwise from the positive x-axis.
The action is derived from the motion performed by the robot.

As the learner's position may not be fixed, \sanet{} seeks to recognize the coordinates and orientation of the observed object(s) relative to the learner's location and project it on the global reference frame. While the RGB frame offers context, the depth data is relative to the observer. Coordinates are recognized by interleaving convolutional neural nets (CNN) and pooling layers followed by fully-connected layers input to a softmax. This allows the use of all four channels, RGB-D, in recognizing the coordinates. Identifying the expert's orientation and action is more challenging. Both of these rely on temporal data, and \sanet{} utilizes frames from time steps $t-2$, $t-1$, and current time step $t$. Each frame is cropped previously by a network such as Faster R-CNN~\cite{DBLP:journals/corr/GirshickDDM13} or YOLO2~\cite{yolov3} to focus attention on the expert. The network backtracks the movement inside the bounding box for time step $t-1$ and $t-2$ using a layer of time-distributed CNNs followed by two convolutional long-short-term memory nets (LSTM)~\cite{Hochreiter97longshort-term}. \sanet{} continues to utilize the depth channel here by running an intercept to the previously described fully-connected nets that provides the relative distance.

We evaluate \sanet{} on two diverse tasks. $(i)$ It is used to identify the state-action sequences of two TurtleBots that are simultaneously but independently patrolling a hallway. \sanet{} is deployed on a third TurtleBot that is observing the patrollers from a vantage point, and is tasked with penetrating the patrol~\cite{bogert2014multi}. 
$(ii)$ \sanet{} is used to identify the state-action sequences of a PhantomX robotic arm that is engaged in pick-and-place to sort objects. In both tasks, \sanet{} exhibits high accuracy while being able to run on computing machines with limited processing power and memory on board a robot. Ablation and robustness studies demonstrate that the architecture is effective and that \sanet{} can handle some typical adverse conditions as well. Consequently, \sanet{} offers high-accuracy trajectory recognition to facilitate robots engaged in LfO for various tasks.

\section{Related Work}
\label{sec:related}

%Recognizing state-action pairs from videos is a fundamental step for learning from demonstration and has been studies for years. 
Traditionally, the state and action of an observed robot is recognized by tracking a marker associated with the robot. For example, Bogert and Doshi~\cite{bogert2014multi} utilizes a colored box on top of the TurtleBot for blob detection in CMVision coupled with 3D point-cloud processing. This centroid-based method simplifies the estimation of the robot's trajectory but is not robust to occlusion of the marker and to noise in the context.

% which simplifies the detection of the robot and estimation of its state and action. A limitation of such methods is a lack of robustness to occlusion of the object and to  noise in the context. 

%There are many options to learn state and action from videos. It can be just a normal video stream or added with depth. One Approach is to use attach some sort or identifier like a colored shape for example a box; \cite{Bogert:2014:MIR:2615731.2615762} to the observed object. Based on the this box the centroid is calculated to get the state and the movement of this box gives us the action. While the results given by this method is very good. The method does not work when there is large amount of noise or due to occlusion.

Recently, deep NNs have demonstrated significantly improved performance on tasks involving image and video analysis~\cite{DBLP:journals/corr/HeZRS15,yue2015beyond}. Related to our method are the NN architectures in computer vision for recognizing human gestures and activities though these rely predominantly on videos. For example, Ji et al.~\cite{6165309} recognizes human actions in surveillance videos using a 3D-CNN. Furthermore, a recurrent NN (RNN) combined with 3D-CNN~\cite{DBLP:journals/corr/MontesSN16}  classifies and temporally localizes activities in untrimmed videos. Rezazadegan et al.~\cite{Rezazadegan17:Action} introduced two-stream VGG16 based CNNs that utilize spatial and optical flow images to recognize robotic activities. 
Depth modality can be leveraged for gesture recognition using two separate CNN streams with a late fusion network~\cite{DBLP:journals/corr/EitelSSRB15}. Alternately, RGB and depth modalities can be superimposed to extract features for action recognition with CNNs~\cite{DBLP:journals/corr/WangLGZTO17}.  These action recognition methods learn from videos by treating them as either 3D volumes with multiple adjacent frames~\cite{6165309}, one or multiple compact images~\cite{park2016combining,Rezazadegan17:Action,DBLP:journals/corr/WangLGZTO17}, or as an image frame sequence~\cite{DBLP:journals/corr/MontesSN16}. Our method belongs to the last category and handles the image and depth sequence with LSTMs, which learn temporal dependencies. 

% \yh{Different from existing methods for action recognition that focus on identifying human activity from videos, our subgoal of action recognition is to recognize the motion from a couple of frames in real time. A fancy network design for action recognition in computer vision probably does not suit the action recognition task in our robotic context. 

Among methods that use LSTMs, Wang et al.~\cite{wang2016beyond} adopts 3D-CNNs to extract features from video clips and uses the LSTM to extract dynamic features for recognizing actions in the video. 
%however, we aim to identify the robot's action from as few frames as possible, e.g., three frames. 
Another related work~\cite{ullah2017action} leverages a bidirectional LSTM to capture temporal changes. But, this may not distinguish between moving forward and backward or left vs. right motions. In comparison to these sophisticated NN designs, \sanet{}'s architecture for action recognition is simpler because it avails of an additional modality -- depth streams. Furthermore, in contrast to these methods focusing on recognizing actions, \sanet{} is tasked with recognizing the state and action pairs simultaneously in real time from just a few frames, as demonstrated by the experiments. 

A recent deep NN architecture SE3-Nets~\cite{byravan2017se3} predicts the rigid body motion of objects in the robotic scene. However, \sanet{} has a different focus: to recognize or summarize a robot's trajectory from RGB-D streams rather than estimate its visual representation using 3D point cloud or images.

\begin{figure*}[t]
\centerline{\includegraphics[width=0.8\textwidth]{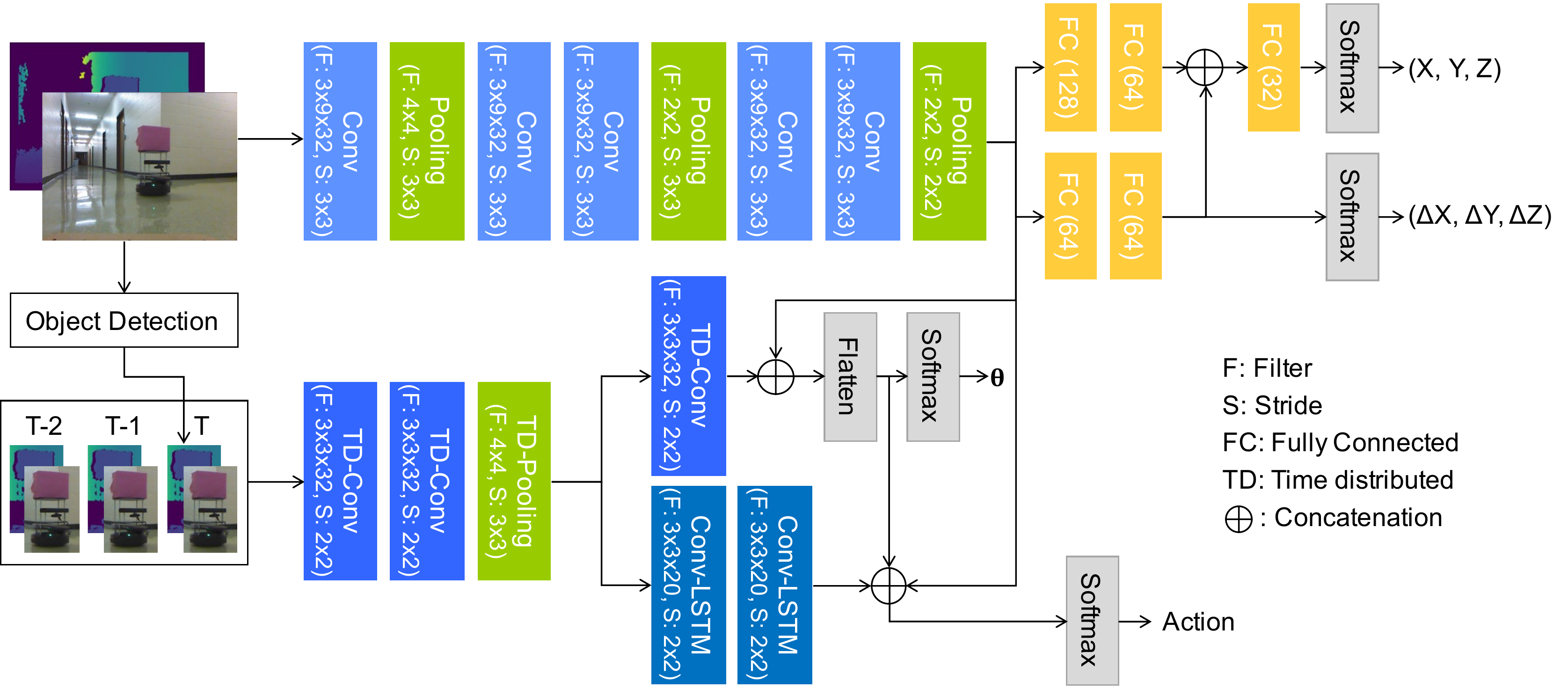}}
\caption{An overview of the \sanet{} architecture. This network jointly predicts the state and action of an expert using the observed RGB-D data streams and corresponding sequential data cropped by an object detection model. The final outputs of the network include the coordinate $X, Y, Z$, the orientation $\theta$, and the action. The additional output of relative coordinate $\Delta X, \Delta Y, \Delta Z$ is used in the training.}
\label{fig:sanet}
\end{figure*}

\section{\sanet{} Architecture}

As \sanet{} is tasked with recognizing state-action pairs, this  motivates a network design that efficiently mixes convolutional and recurrent NNs, which we describe below. 

\subsection{Problem Definition}

We aim to automatically estimate the state and action pairs of an expert from RGB-D streams using deep NNs. Given the expert's three video frames captured by a learner at time points $t-2$, $t-1$, and $t$, our network jointly predicts the state ($X, Y, Z, \theta$) and action ($A$) of the expert at the current time point $t$. Here, the tuple ($X, Y, Z$) in the state representation describes the location coordinate of the expert in a 3D environment; the $Z$ dimension is ignored for 2D cases. The $\theta$ describes the orientation of the expert. In this paper, we consider discrete state and action spaces, which allows formulating the task as a multi-label classification problem. Formally, the problem can be formulated as:
\begin{small}
\begin{align*}
    (&X, Y, Z, \theta, A) = f(I_{t-2}, I_{t-1}, I_t; \bf{\Theta}), ~~~~\text{where}\\
    &X \in \{0, ..., N_X-1\}, Y \in \{0, ..., N_Y-1\}, Z \in \{0, ..., N_Z-1\}, \notag \\
    &\theta  \in \{0, ..., N_\theta-1\}, 
    A \in \{0, ..., N_A-1\}. \notag
\end{align*}
\end{small}
Here, $f$ denotes the mapping function learned by our classification network; $I_{t-2}$, $I_{t-1}$, and $I_t$ are the three frame inputs; $\bf{\Theta}$ represents the parameter set of the network for classifying the state and action jointly; $N_X$, $N_Y$, and $N_Z$ are the discretized dimensions in each coordinate; $N_\theta$ is the number of the expert's orientations -- for instance, we have four orientations including north, south, east, and west in the TurtleBot application; $N_A$ is the number of actions, e.g., four actions including move forward, stop, turn right, and left. Overall, the network includes two coupled components for the state and action recognition, which are learned simultaneously. \sanet{}'s architecture is shown in Fig.~\ref{fig:sanet}. 

\subsection{State Recognition}

\begin{figure}[!ht]
\centering
\begin{tikzpicture}[thick, spy using outlines={rectangle,lens={scale=2.5}, width=1cm, height=1.5cm, connect spies}]
	\node (reg_id1) {\includegraphics[width=\columnwidth]{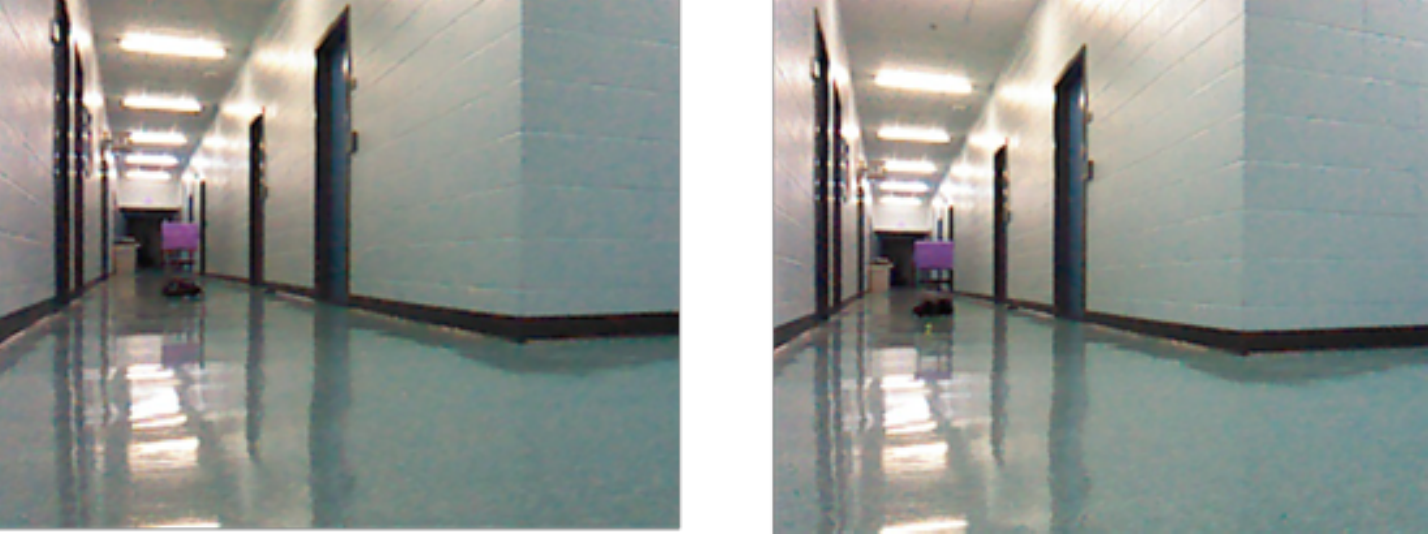}};
 	\spy [red] on (-3.2, 0.05) in node (lspy) [left,fill=black!5] at (-0.5,0.6);
	\spy [red] on (1.35,-0.05) in node (lspy) [left,fill=black!5] at (3.15,0.6);
\end{tikzpicture}
\caption{An example of a TurtleBot in two similar images but having different orientations.}
\label{fig:orientation_example}
\end{figure}

State recognition aims to determine the expert's coordinate ($X, Y, Z$) and its orientation $\theta$. Typically, the expert's coordinate can be identified on the basis of its surrounding environment. Therefore, we use one image frame without considering the temporal information in our coordinate recognition module. Different from state recognition, orientation recognition requires more than one image frame to recognize hard-to-distinguish orientations. As shown in Fig.~\ref{fig:orientation_example}, the TurtleBot is oriented differently in the two images, but the image difference is too subtle to correctly separate these two orientations of the TurtleBot. In such situations, image sequence plays an important role in recognizing the orientation. Therefore, in the state recognition of \sanet{}, we separate the prediction of the coordinate ($X, Y, Z$) from that of the orientation $\theta$, as one network stream takes the static image as input while the other takes the image sequence.

\noindent \textbf{Coordinate recognition}
As shown in the top stream of the network in Fig.~\ref{fig:sanet}, only the image frame at time point $t$ is used to predict the expert's location coordinate. 
%Figure~\ref{fig:domains} shows the discretized states in both 2D and 3D grids. 
We assign a pre-defined coordinate system for each environment; that is, each image frame will be classified into a unique coordinate, which is represented by an absolute location ($X, Y, Z$) with respect to the origin in the coordinate system. The expert's coordinates are learned from images captured by the learner; however, the learner's location may change in different situations. To improve the network's generalization, we leverage the relative distance between the expert and the learner to help in the recognition of the expert's coordinate. 

In the coordinate recognition branch, we have two sets of coordinate-related predictions: the relative distance ($\Delta X, \Delta Y, \Delta Z$) and the absolute coordinate ($X, Y, Z$). These two prediction tasks share the same process of image feature extraction, which includes five convolutional layers and three max pooling layers. The convolutional layers use 32 filters with the same kernel size $3 \times 9$ and the same $3 \times 3$ stride. The three pooling layers are located after the first, third, fifth convolutional layers, respectively, with filters of size $4 \times 4$, $2 \times 2$, and $2 \times 2$ and strides of size $3 \times 3$, $3 \times 3$, and $2 \times 2$. Following the convolutional and pooling layers, two fully convolutional (FC) layers are used in the classification. As the prediction of relative distance contributes to coordinate prediction, we have an additional FC in the stream of coordinate classification after concatenating the pre-activation of the softmax from the relative distance classification. 

\noindent \textbf{Orientation recognition} Different from the coordinate parameter, the orientation of the expert guides its movement particularly so in the context of mobile robots.
%-- similar to the action discussed later in Section~\ref{sec:action}. 
Therefore, in both orientation and action recognition we would like the network to have its attention on the expert itself, especially when the expert is far away from the learner and relatively small in the whole image frame. To achieve this goal, we adopt object detection to make the expert stand out for perceiving its behavior. More details about the object detection are given in Section~\ref{sec:detection}. After object detection, we have three new sequential frames, which are cropped from the original RGB-D image inputs and re-sized to images of size $150\times 100$ to facilitate orientation and action recognition of the expert. The sequential frames are essential in orientation recognition to differentiate hard examples as shown in Fig.~\ref{fig:orientation_example}.

To handle the sequential image inputs, we use time-distributed convolutional (TD-Conv) layers in the stream for recognizing orientation. These layers collect image features required for orientation recognition from all three time steps. In particular, we have two TD-Conv layers, followed by one time-distributed max pooling layer and another TD-Conv layer. Each TD-Conv layer has 32 filters of size $3\times 3$ and stride of $2\times 2$, and the pooling layer uses a filter of size $4\times 4$ and stride of $3\times 3$. We observe that the orientation and action recognitions are connected to coordinate recognition, albeit loosely. To motivate this, note that the TurtleBot is less likely to turn left or right if it is in the middle of a corridor. Thus, we concatenate the whole-image features extracted from the coordinate recognition with the spatio-temporal features extracted from the cropped image sequence to predict the expert's orientation. A similar operation is performed in  action recognition, as discussed next. 

\subsection{Action Recognition}
\label{sec:action}

Similar to the orientation recognition, actions are recognized using  the same three sequential cropped images after object detection (Section~\ref{sec:detection}). The goal is to determine the expert's action -- for example, in which cardinal direction is the expert moving. Because the orientation and action recognition are working on the same input, they share the first three layers for extracting lower-level features from cropped images at all time steps. The action recognition then uses two convolutional LSTM layers to further compose higher-level features and capture temporal changes in the image sequence. These two new layers also use 32 kernels of size $3\times 3$ and stride of $2\times 2$. In this branch we leverage all features extracted from the state (both coordinate and orientation) recognition to support the action recognition. In tasks involving mobile robots, the orientation and action are often coupled and we use concatenation to make full use of extracted features. Orientation helps in predicting the action but is, of course, not sufficient. A mobile robot moving toward the observer exhibits the same orientation, which does not reveal the move-forward action. If the state and action are known to be independent, these connections would be removed and \sanet{} handles the two independently.     

%\begin{figure}[t]
%\centerline{\includegraphics[width=0.45\columnwidth]{Figs/crop_factor.eps}}
%\centerline{\includegraphics[width=0.45\columnwidth]{Figs/crop_factor_agg.eps}}
%\caption{Overview.}
%\label{fig:object_detection}
%\end{figure}

\subsection{Expert Detection}
\label{sec:detection}

This object detection module provides inputs for the orientation and action recognition of the expert. We use the RGB data stream at the current time point $t$ to perform the object detection using an existing model, YOLO2~\cite{yolov3}. Using the predicted bounding box for the expert $[(x_1, y_1), (x_2, y_2)]$, we crop the images from the frames $t-2$, $t-1$, and $t$. We keep a small amount of surrounding environment background; this buffered cropping is calculated by the linear equation,
$\Delta_a = r \times \Delta_b + c_{min}$.
% \begin{equation*}
% \Delta_a = r \times \Delta_b + c_{min}.
% \end{equation*}
Here, $r \geq 1$ is the cropping factor that determines how aggressively the users want to crop the image; $\Delta_b$ is the width $|x_1 - x_2|$ or the height $|y_1 - y_2|$ of the bounding box before the buffered cropping, while $\Delta_a$ is the corresponding value after the cropping; and $c_{min}$ is the minimum amount of cropping, e.g., 10 pixels. In all of our experiments, we set $r=1.1$.

\subsection{Masking for Learning from Multiple Experts}
\label{sec:masking}

In practice, we may have more than one expert. However, \sanet{} is not explicitly designed to recognize the state and action pairs of multiple experts. Despite this, we can allow multiple experts by using masking that ensures only one expert exists in the images for recognition. Specifically, we leverage the object detection (Section~\ref{sec:detection}) to separate the experts and generate a new image for each. To generate the image for one expert, we remove all others  using their detected bounding boxes and replace the removed regions with the background image stored in memory. Thus, we have new images for each expert to pass through the net for state-action recognition. As shown later, this use of masking to segregate experts in each frame does not negatively impact the simultaneous tracking of multiple robots across frames. 

\section{Experiments}
\label{sec:experiments}
%----------------------------------------------------------------------

\sanet{} exhibits a general architecture useful in multiple domains. We evaluate it on two tasks offline and online on a physical robot and report on our extensive experiments.

\subsection{Tasks}

\begin{figure}[!ht]
\begin{minipage}{4.5in}
\centerline{\includegraphics[height=2.0in]{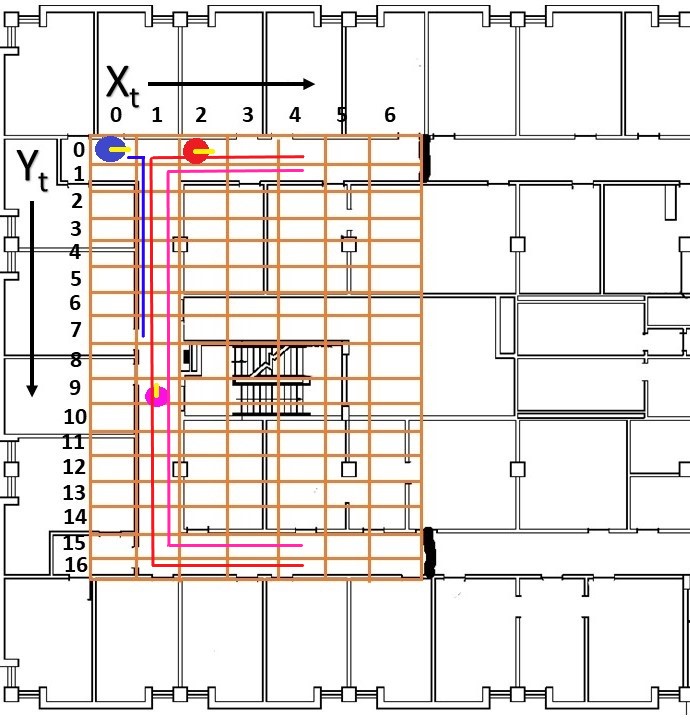}}
\centerline{\small{$(a)$}}
\end{minipage}
\begin{minipage}{4.5in}
\centerline{\includegraphics[height=1.5in]{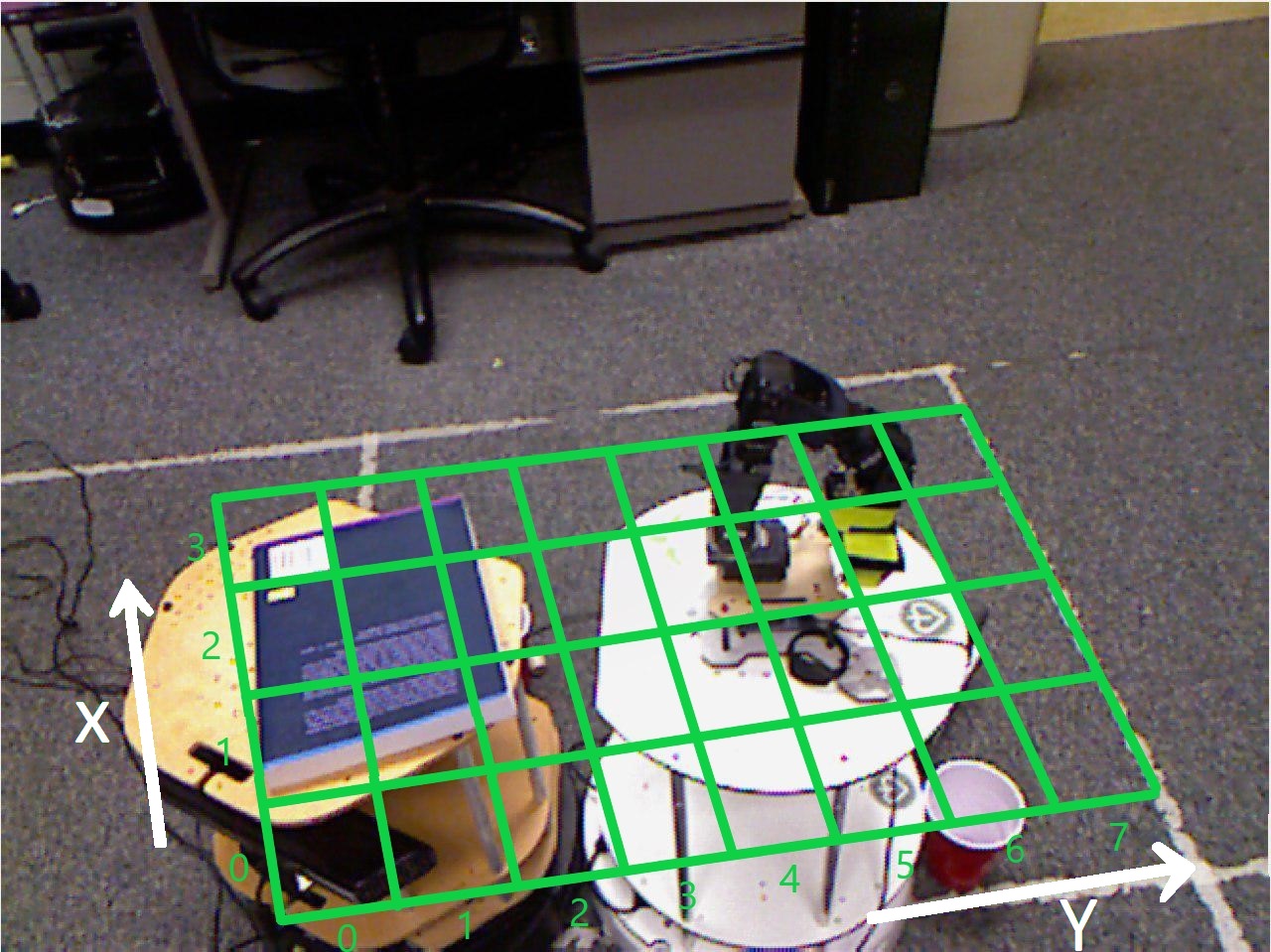}
\hspace{0.2in}
\includegraphics[height=1.5in]{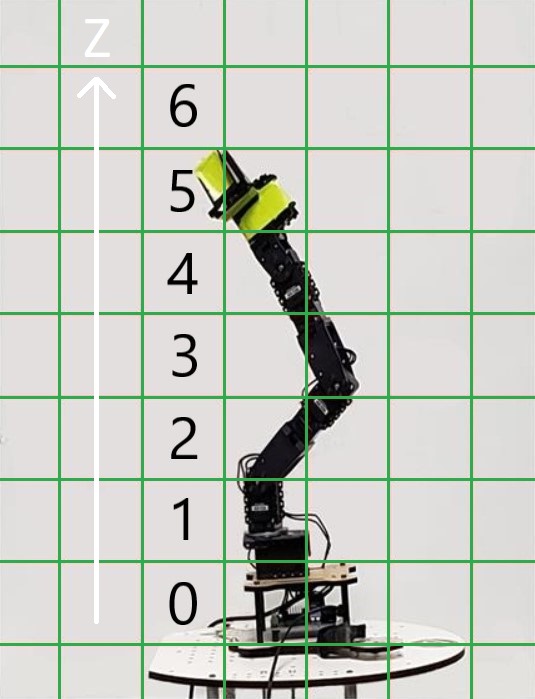}} 
\centerline{\small{$(b)$}}
\end{minipage}
\caption{$(a)$ A map of the hallway patrolled by two TurtleBots. The learner, shown in blue, observes the patrols from its vantage point using a Kinect 360 RGB-D sensor. A 2D grid is superimposed on the hallways. $(b)$ \sanet{} is deployed on a computer connected to a Kinect 360 that observes a PhantomX arm mounted on a TurtleBot. A 3D grid is superimposed for the coordinates of its end effector.}
\label{fig:domains}
\end{figure}

We evaluated \sanet{} on two diverse LfO tasks. First, it was deployed on a TurtleBot tasked with penetrating cyclic patrols by two other TurtleBots in a hallway as shown in Fig.~\ref{fig:domains}$(a)$. Each patroller can assume one of 4 orientations and 4 actions. This task replicates a well-known testbed for evaluating  LfO methods such as inverse reinforcement learning~\cite{bogert2014multi,Bogert15:Toward}. The other task involves observing a PhantomX arm mounted on a TurtleBot (Fig.~\ref{fig:domains}$(b)$) and performing  pick-and-place to sort objects of two types.
An overlooking Kinect 360 RGB-D sensor observes the arm~\cite{Trivedi18:Inverse}. This task adds a third dimension, the height of the end effector, to the state, and the arm has 6 possible orientations and 6 actions that correspond to the motion of the end effector in the 3D space.

\subsection{Formative Evaluation}

We evaluated \sanet{} using stratified 5-fold cross validation for both tasks. 500 RGB and depth image pairs, each annotated with a bounding box only, were utilized to train a Faster R-CNN~\cite{DBLP:journals/corr/RenHG015} whose output then trained a YOLO2 net to learn the bounding boxes for cropping the images. The complete data had 60K annotated sets of RGB and depth image frame pairs for the patrolling task and 10K  sets for the manipulation task. Each set consists of an uncropped pair and three cropped pairs at time points $t-2$, $t-1$, and $t$.

%The cropped and uncropped RGB-D frames were both provided as input to the top and bottom streams of \sanet{}, respectively.

\begin{table*}
\setlength{\tabcolsep 2pt}
    \centering
    \small
    \begin{tabular}{llccccc}
        \toprule
        Task & Method & X & Y & Z & $\theta$ & Action  \\
        \midrule
        %Run 1 & 98.853 & 99.96 & 99.99 & 99.97\\
        %Run 2 & 98.853 & 99.96 & 99.99 & 99.95\\
        %Run 3  & 98.853 & 99.95 & 100 & 99.95 \\
        %Run 4  & 98.885 & 99.97 & 99.97 & 99.94 \\
        %Run 5  & 98.81 & 99.99 & 100 & 99.97 \\
        \multirow{2}{*}{\bf Patrolling} & \sanet{} & 98.85$\pm$0.02 & 99.97$\pm$0.014 & --- & 99.99$\pm$0.01 & {\bf 99.74$\pm$0.01} \\
        & Two-Stream CNN~\cite{Rezazadegan17:Action}& --- & --- & --- & --- &87.53$\pm$0.61\\ 
    %     \bottomrule\\
    % \end{tabular}
    % \begin{tabular}{llccccc}
    %         \toprule
    %         Domain & Method & X & Y & Z & $\theta$ & Action  \\
            \midrule
            %Run 1 & 97.63 & 95.23 & 96.54 & 98.19 & 99.12\\
            %Run 2 & 97.65 & 95.19 & 96.56 & 98.12 & 99.14\\
            %Run 3  & 97.62 & 95.2 & 96.58 & 98.23 & 99.1\\
            %Run 4  & 97.63 & 95.22 & 95.59 & 98.17 & 99.14\\
            %Run 5  &97.66 & 95.21 & 95.55 & 98.15 & 99.16\\
            \multirow{2}{*}{\bf Manipulation} & \sanet{} & 97.64$\pm$0.02 & 95.22$\pm$0.01 & 96.16$\pm$0.49 & 98.17$\pm$0.04 & {\bf 99.13$\pm$0.02}\\
            & Two-Stream CNN~\cite{Rezazadegan17:Action}& --- & --- & --- & --- & 84.05$\pm$3.39\\
            \bottomrule
    \end{tabular}
    \caption{Mean and standard deviation of \sanet{}'s performance from a 5-fold cross validation for the patrolling and manipulation tasks. We show the prediction accuracy of state and action for both tasks. This is compared with another deep NN based method that performs action recognition only from spatial and optical flows~\cite{Rezazadegan17:Action}. Note that ` --- ' denotes not applicable.}
    \label{tab:sanet-results}
\end{table*}

\begin{table*}[!ht]
    \centering
    \small
        \begin{tabular}{lcccc}
            \toprule
            Ablation & X & Y & $\theta$ & Action  \\
            \midrule
                \sanet{} w/o Relative X and Y & {\bf 81.38$\pm$1.50} & 91.44$\pm$1.46 & 91.24$\pm$1.26 & {\bf 83.63$\pm$1.86} \\
                \sanet{} w/o data from $t-1$, $t-2$ & 96.56$\pm$0.01 & 98.32$\pm$0.01 & {\bf 79.43$\pm$1.33} & {\bf 78.86$\pm$3.57} \\
                \sanet{} w/o depth channel & {\bf 87.23$\pm$1.50} & 95.12$\pm$1.49 & {\bf 83.56$\pm$1.52} & {\bf 81.12$\pm$1.00}\\
                \sanet{} w/o object detect & {\bf 68.74$\pm$2.26} & {\bf 69.95$\pm$1.26} & {\bf 21.65$\pm$2.39} & {\bf 33.89$\pm$0.96} \\
            \bottomrule
        \end{tabular}
    \caption{Four ablation experiments in the patrolling task and the impact on the prediction accuracy of state and action. }
    \label{tab:ablation}
\end{table*}

Table~\ref{tab:sanet-results} shows the prediction accuracy on the 2D or 3D coordinates and orientation that make up the state, and on the action for each task. We show the mean and standard deviation across the 5 runs. Notice that in both problems, \sanet{} generates predictions of state and action with very high accuracy, with those for the manipulation task being slightly less accurate than those for the patrolling. This is generally consistent across all folds yielding low standard deviations. The action recognition compares favorably to that by the two-stream CNN~\cite{Rezazadegan17:Action} previously discussed in Section~\ref{sec:related}.

\subsection{Ablation Study}

We performed an ablation study on a smaller data set to understand the sensitivity of \sanet{}'s performance on its key components. The ablation study removes a part of the network and conducts experiments on the revised model.

\noindent \textbf{Relative X and Y}~ In this experiment on the patrolling task, we eliminate that part of \sanet{} which contributes to establishing the 2D grid coordinate of the observed robot relative to the observer's location. This part relies more on the depth data. Consequently, we may expect the network to memorize the location by relying more on RGB data but unable to detect changes in its own deployed position.  %There over fitting on training set with almost 100\% results but when tested on training set we get the results on table \ref{tab:abalation x and y}
Row 1 of Table ~\ref{tab:ablation} shows a significant drop in the prediction accuracy of state and action with a more pronounced drop in the accuracy of predicting the X-coordinate and action. These two rely significantly more on the relative distances.

\noindent \textbf{Temporal sequence data}~ In this experiment, we eliminate the part of \sanet{} responsible for processing temporal data from previous time steps $t-1$ and $t-2$. This also eliminates those two input channels and keeps input from time step $t$ only. We hypothesize this removal to significantly impact the recognition of orientation $\theta$ and action, both of which are thought to rely on sequence data. 
%~\footnote{To illustrate, it is difficult to identify the orientation of a man wearing a face mask on his back unless he starts walking.} 
On the other hand, a single frame could be sufficient to identify the orientation in many cases. 
% nformation  $\theta$ and check if $\theta$ really does need the temporal information. Since we as human beings can recognize the direction of an object from a single image. The network should be able to recognize the direction ($\theta$) without the use of temporal data. Since theta($\theta$) is required to understand the action the accuracy of action will reduce.
Table~\ref{tab:ablation}, row 2 presents prediction accuracies that are significantly lower for $\theta$ and action, while recognizing the 2D coordinates is generally not affected. As such, the temporal data is indeed important for \sanet{} in general.

\noindent \textbf{Multimodal data}~ Next, we study if depth data is needed for the predictions and how the network will behave when its removed. Can we make the network learn the state and action from RGB data only? Row 3 of Table~\ref{tab:ablation} shows that the predictions of X-coordinate, $\theta$, and action are significantly degraded in the absence of the depth channel. The Y-coordinate is least impacted as we may expect. As a patroller approaches the observer, there are multiple states for which the RGB frames are similar. In the absence of depth, the network memorizes certain features and overfits on those. 

\noindent \textbf{Object detection}~ Finally, we precluded the object recognition performed by YOLO2, resulting in no cropped images as input. 
%; \sanet{} did not take any input as cropped images. 
The drastic drop in prediction quality of all coordinates, orientation, and action (row 4) gives evidence that object detection is required. Coordinate recognition suffers because object detection is needed for masking each expert in the context of multiple experts. In recognizing the orientation and action, object detection plays a more integral role focusing \sanet{}'s attention, which is demonstrated by a larger degradation in their prediction accuracy.

\begin{table*}[!ht]
    \setlength{\tabcolsep 4pt}
    \centering
    \small
        \begin{tabular}{llccccc}
            \toprule
            Task & Method & X & Y & Z & $\theta$ & Action  \\
            \midrule
            \multirow{2}{*}{\bf Patrolling} & \sanet{} & {\bf 97.23$\pm$0.29} & {\bf 98.12$\pm$0.49} & --- & {\bf 96.25$\pm$0.67} & {\bf 96.16$\pm$0.68}\\
            
            & Centroid method~\cite{bogert2014multi} & 94.15$\pm$0.00 & 96.13$\pm$0.00 & --- & 93.16$\pm$0.00 & 78.26$\pm$0.68\\
            
            \midrule
            
            {\bf Manipulation} & \sanet{} & 87.56$\pm$0.02 & 89.25$\pm$0.02 & 91.12$\pm$0.03 & 88.32$\pm$0.01 & 91.18$\pm$0.01\\
            \bottomrule
        \end{tabular}
    \caption{\sanet{}'s accuracy in physical experiments for the two tasks under {\em typical conditions}. Action prediction is much improved over the baseline method for the TurtleBots.} %The z-coordinate does not apply for the ground robots.}
    \label{tab:results_physical}
\end{table*}

\begin{table*}[!ht]
    \centering
    \small
        \begin{tabular}{lccccc}
            \toprule
            Test & X & Y & $\theta$ & Action  \\
            \midrule
            \sanet{} w/ Noise & {\bf 92.65$\pm$0.87} & {\bf 96.65$\pm$0.72} & {\bf 95.23$\pm$0.40} & {\bf 95.12$\pm$0.76} \\
            Centroid method w/ Noise & 34.20$\pm$6.62  & 44.43$\pm$1.42 & 23.23$\pm$1.31 & 42.45$\pm$1.88\\
            \midrule    
            \sanet{} w/ Occlusion  & {\bf 45.15$\pm$0.87} & {\bf 54.60$\pm$0.76} & {\bf 64.12$\pm$0.99} & {\bf 46.36$\pm$1.00} \\
            Centroid method w/ Occlusion & 18.23$\pm$2.13 & 17.34$\pm$1.57 & 14.42$\pm$0.15 & 43.12$\pm$0.80 \\
            \bottomrule
        \end{tabular}
    \caption{Robustness testing of \sanet{} and the centroid method~\cite{bogert2014multi} on the patrolling task with background noise and occlusion.}
    \label{tab:robustness}
    \end{table*}

\subsection{Summative Evaluation on Physical Robots}

\begin{figure}[!ht]
\begin{minipage}{4.5in}
    \centerline{\includegraphics[width=2.5in]{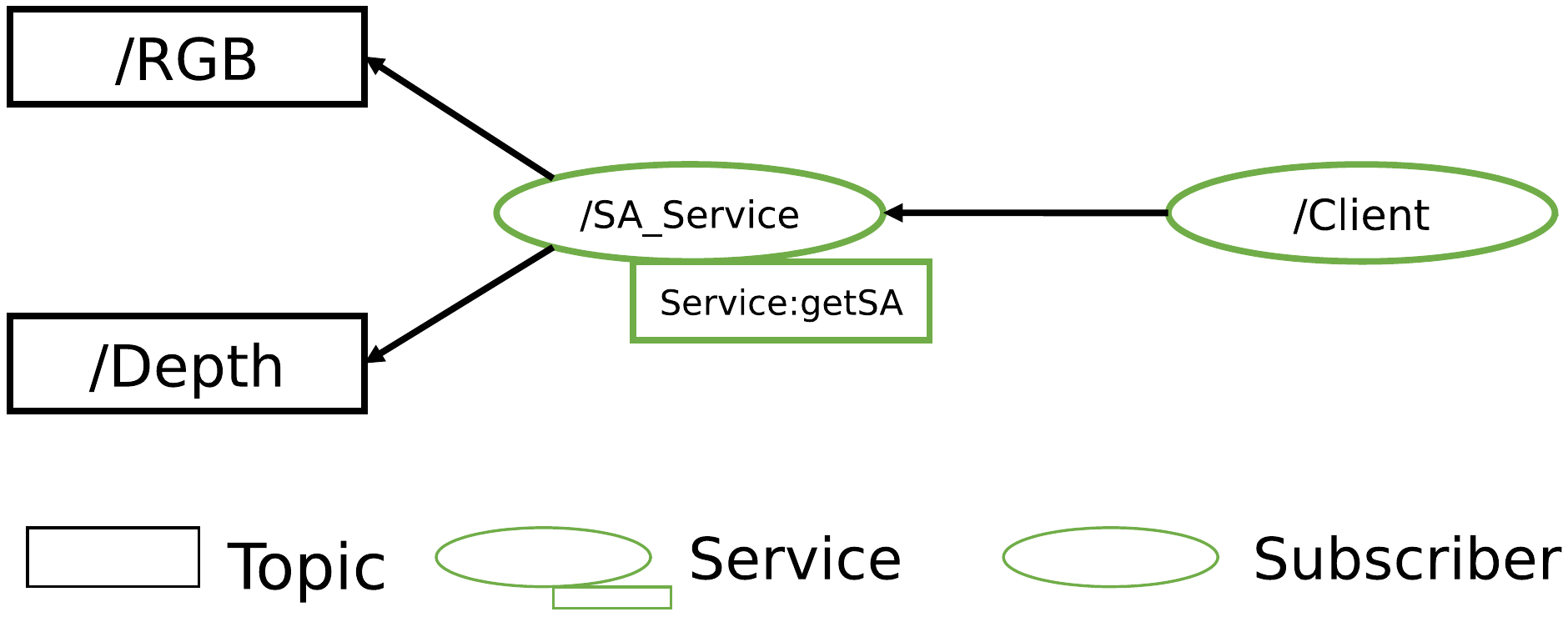}}
    \centerline{\small $(a)$}
\end{minipage}
\begin{minipage}{4.5in}
\small
\center
    \begin{tabular}{lr}
	    \toprule
        Memory usage & 742MB$\pm$ 3MB \\
        \midrule
        Faster R-CNN $\rightarrow$ \sanet{} & 6s$\pm$ 0.4s \\
            YOLO2 $\rightarrow$ \sanet{} & {\bf 1.1s$\pm$ 0.3s} \\
        \bottomrule
    \end{tabular}
    \centerline{\small $(b)$}
\end{minipage}
\caption{$(a)$ ROS nodes architecture for \sanet{} on a robot. $(b)$ \sanet{} resource utilizations on a TurtleBot2 standard ASUS notebook with Intel Core i3, 4GB RAM. Note the run time benefit of YOLO2.}
\label{fig:ROS}
\end{figure}

We deployed the trained \sanet{} with masking (Section~\ref{sec:masking}) on a physical TurtleBot that observed two other TurtleBots patrolling the hallway and on a TurtleBot that is connected to a Kinect 360 overlooking a PhantomX arm picking and placing small objects. \sanet{} can be used in ROS as a service and the corresponding component architecture is shown in Fig.~\ref{fig:ROS}$(a)$.

\begin{figure}[!ht]
\begin{center}
\begin{minipage}{4.5in}
\includegraphics[width=1.0in]{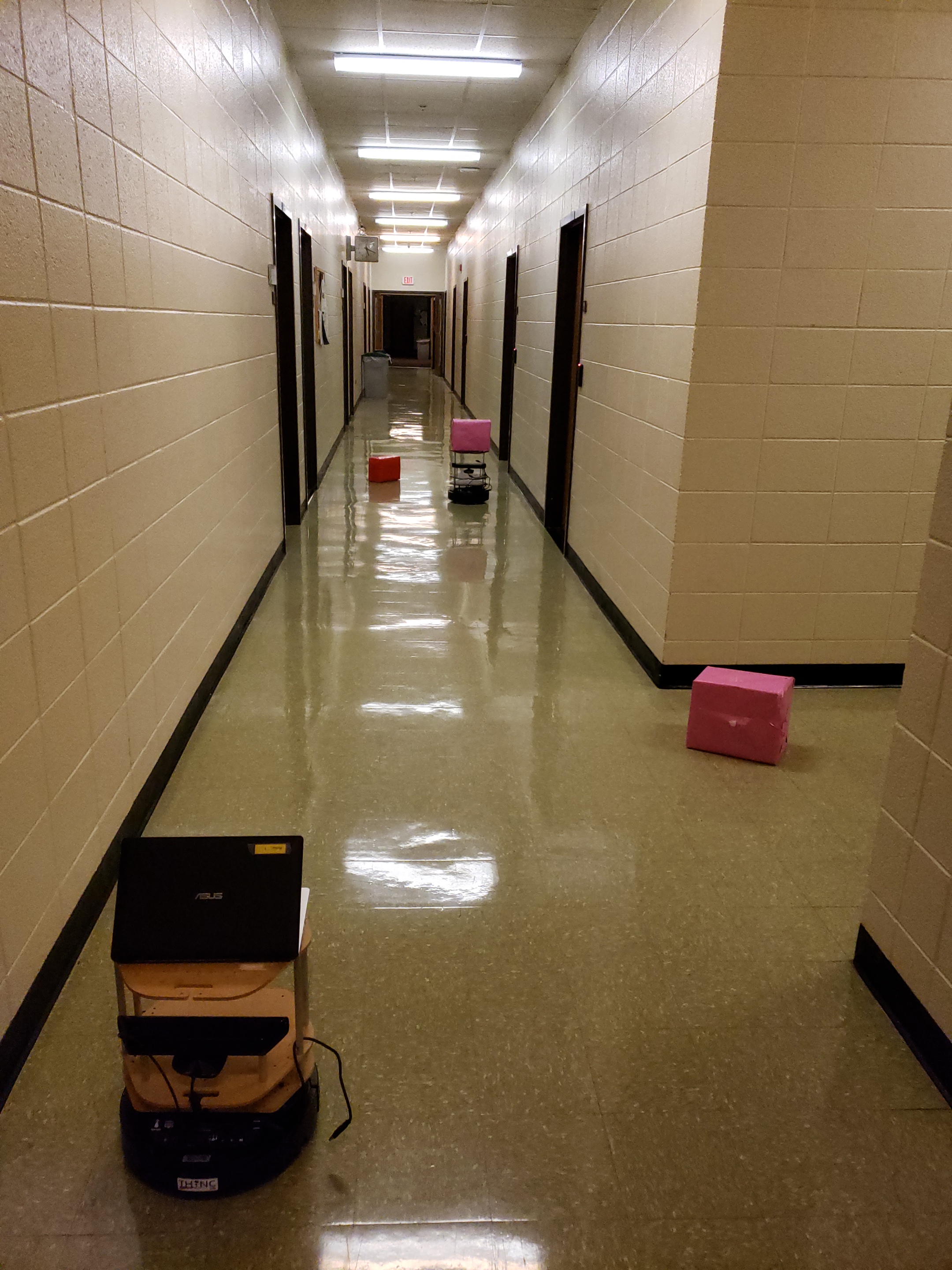}
\includegraphics[width=1.0in]{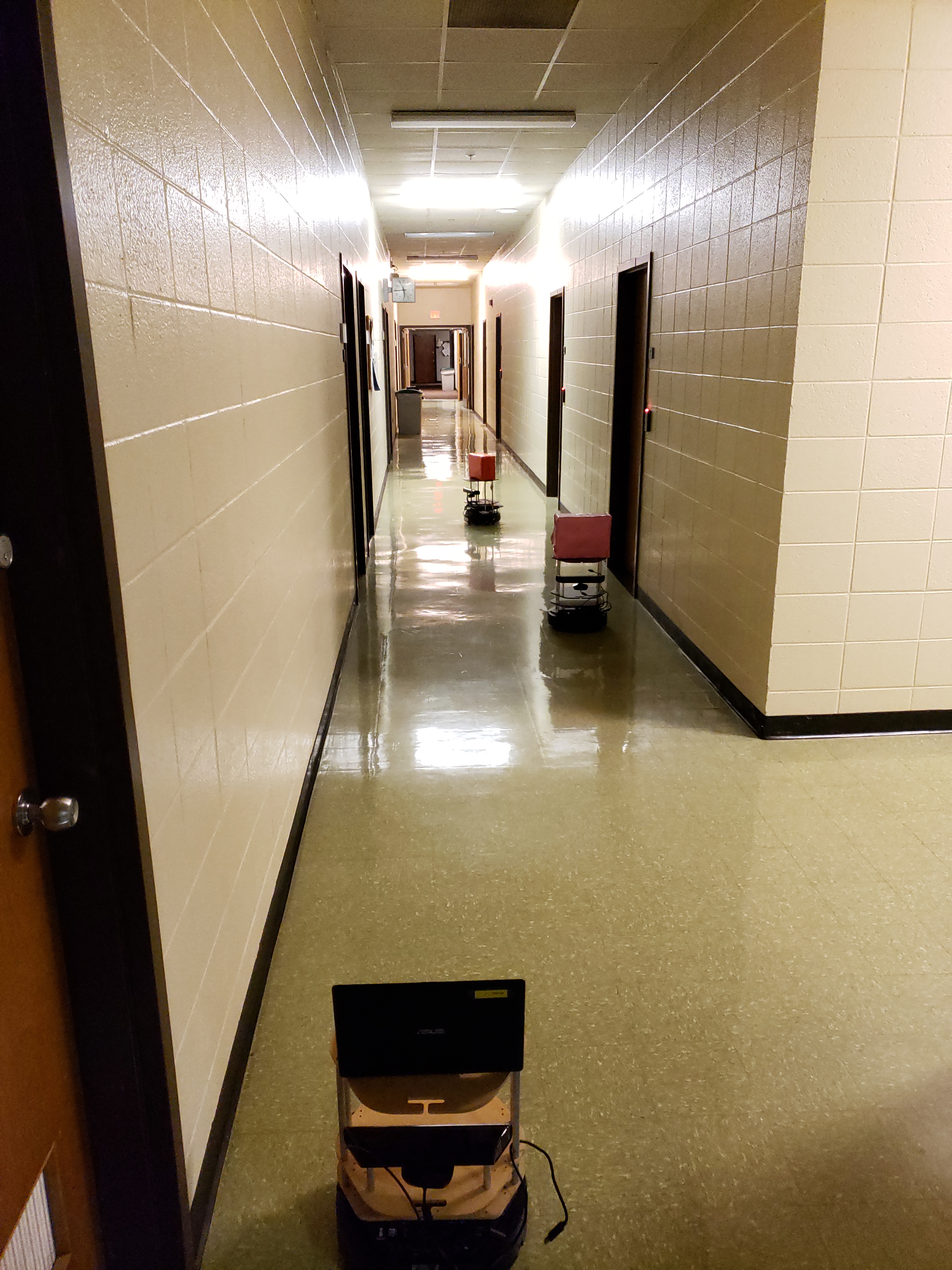}
\includegraphics[width=1.0in]{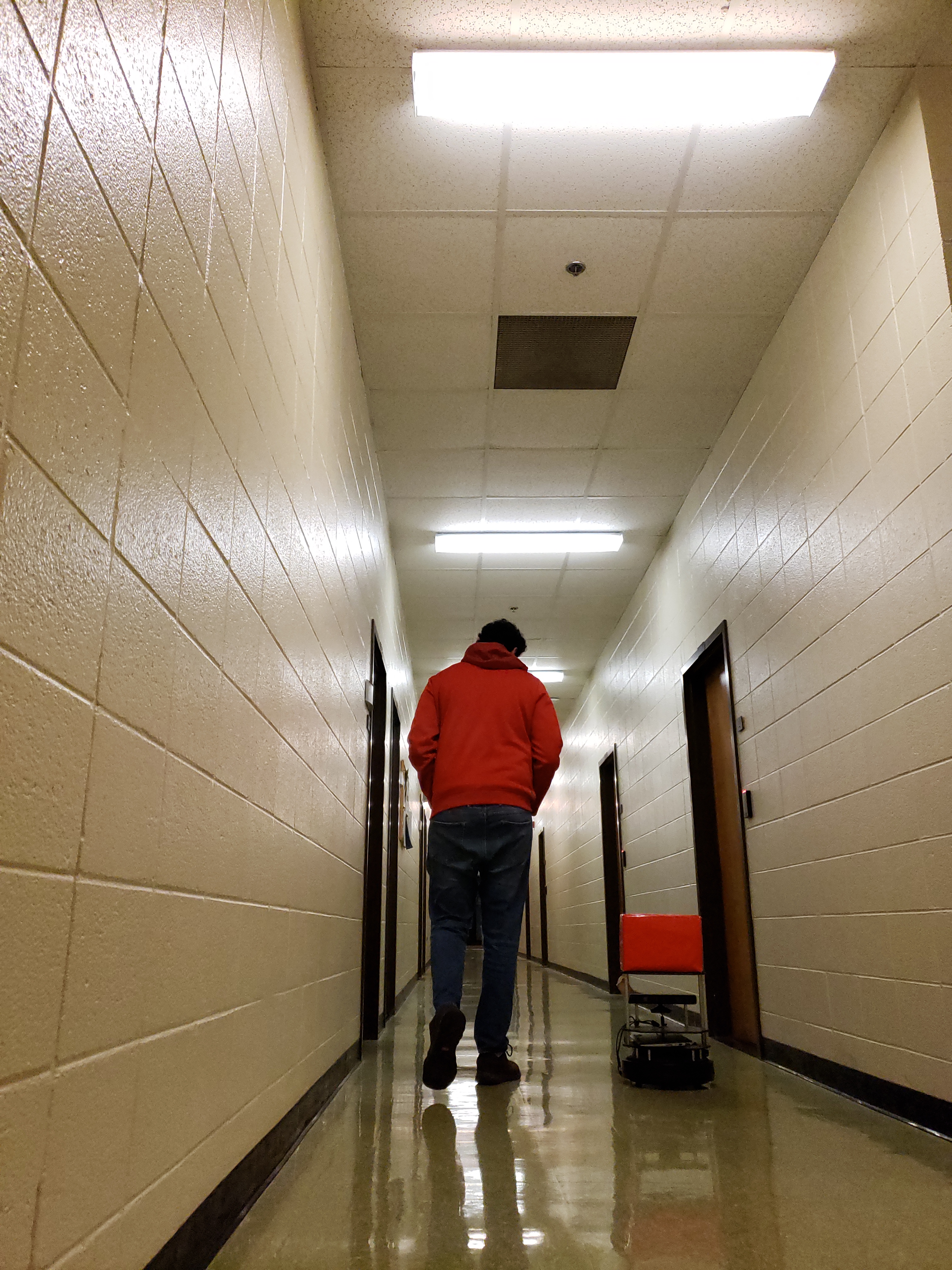}
\includegraphics[width=1.0in]{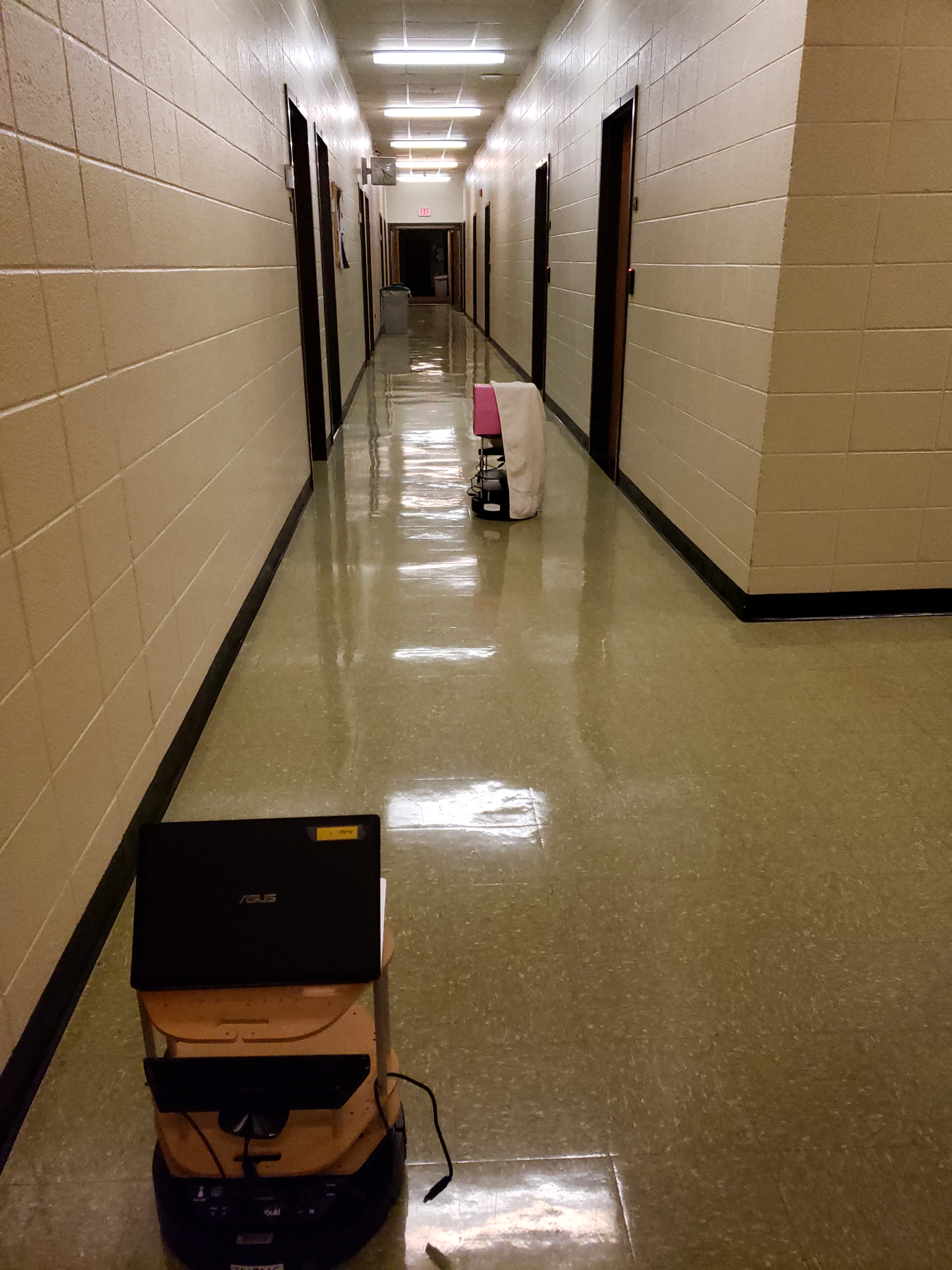}
\end{minipage}
\end{center}
% \begin{minipage}{3.5in}
% \centerline{\includegraphics[width=1.0in]{Figs/occlusion_1.jpg}} 
% \end{minipage}
\caption{Robustness tests involving background noise via similarly colored boxes on the floor, low ambient light, human sharing the space, and observed robot is partially occluded.}
\label{fig:robustness}
\end{figure}

Though, it is generally challenging to report the prediction accuracy in online physical experiments, we logged the RGB-D stream and \sanet{}'s predictions for each frame in the stream. These predictions were later verified manually. Table~\ref{tab:results_physical} reports the prediction accuracy on the observed state-action pairs. We compared \sanet{}'s performance on the patrolling task with a traditional CMVision based implementation that detects the centroid of the colored box on each robot in the Lab color space and analyzes the depth data. Notice that this method, which was utilized previously for this task~\cite{bogert2014multi}, is particularly poor in recognizing the patroller's action and \sanet{} improves on it drastically. As such, \sanet{} should lead to improved LfO. \sanet{}'s reduced accuracy in the manipulation task is due to the increased complexity from a third coordinate and more manipulator actions.
%Since the manipulator arm works on a 3-Dimensional space the only change that was done on the network was changing the last softmax layers to Incorporate the extra Z coordinate and the up and down actions explained in Section \ref{picnplace}. Table \ref{tab:results normal} shows the results performed by our network on the manipulator arm. 
Next, we evaluate \sanet{}'s predictions  under various conditions:

\noindent \textbf{Noise test} In this experiment, we test if background noise impacts the prediction accuracy of the network. The noise is defined as objects that look like or have similar characteristics as the target, and dimmed ambient light. Such background objects, shown in Fig.~\ref{fig:robustness}$(a)$, include a human wearing a similar-colored shirt and boxes of same color. 

 \noindent \textbf{Occlusion test} In this experiment, the target is covered partially to approximate 50\% occlusion; we cover the TurtleBot by a cardboard box or a white cloth as shown in Fig.~\ref{fig:robustness}$(b)$. These robots then patrol the hallways as before. 
 
In Table~\ref{tab:robustness}, we show \sanet{}'s prediction accuracy in each context. For the noise test, the predictions are average of 15 runs split into 5 with a human, 5 with boxes, and 5 with dimmed ambient light. For the occlusion test, again an average of 15 runs is shown with the object partially covered to approximate 50\% occlusion. Notice that \sanet{}'s predictions degrade and rather dramatically under occlusion of the target object. The latter drop is because of \sanet{}'s reliance on RGB data, which get curtailed under occlusion. Nevertheless, it's predictions remain significantly better in both tests than the traditional centroid-based blob detection method. In particular, the centroid-based method fails to detect the observed robots under occlusion. Similarly colored boxes do not excessively impact \sanet{} demonstrating that the  object detection is not critically dependent on the marker.

How much memory is consumed by the ROS deployment of \sanet{}? Figure~\ref{fig:ROS}$(b)$ reports the total amount of RAM held by the ROS service for good performance on state-action recognition.  We also show the maximum time in seconds taken by \sanet{} for prediction when paired with Faster R-CNN and paired with YOLO2 in the patrolling task having two targets. Notice that pairing with YOLO2 speeds up the prediction by a factor greater than five.

\subsection{Evaluating \sanet{}'s Performance on Inverse RL} Finally, we investigate the benefit of using \sanet{} instead of the traditional Centroid method for recognizing state-action pairs in the perimeter patrol domain with the physical TurtleBots. To best evaluate \sanet{}'s effect on the LfO algorithms, here we utilize an online inverse RL method~\cite{arora2019online} as the expert's policy learner. We compare the success rates of penetrating the patrol when using \sanet{} and when using the Centroid method. We performed a total of 30 experiments, involving 3 sets of 10, for each of the two state-action recognition algorithms. In each experiment, the learner observes each of the patrolling robots for exactly 10 trajectories (each trajectory includes 5 state-action pairs), which is then input to the IRL method that learns the patroller's preferences and predicts its trajectory. 
%These predictions are used by the learner's MDP to learn a value function that determines when to start moving toward the goal without being spotted by any of the patrollers. If the attacker is out of the sight of the patrollers until reaching the target point, we count it as a success; otherwise, we count it as a failure. 
After performing 3 sets of 10 runs for each of the two algorithms, the average success rate of online IRL with the Centroid-based technique is 43.9\% with the standard deviation of 7.2\%; the average success rate of online IRL with \sanet{} is 63.3\% with a standard deviation of 8.2\%. Therefore, we see a direct improvement from utilizing \sanet{} toward LfO in a well-known task that has been utilized previously for evaluating inverse RL methods.

%-----------------------------------------------------------------
\section{Concluding Remarks}
\label{sec:conclusion}
%-----------------------------------------------------------------

\sanet{} brings the recent advances in deep supervised learning to bear on a crucial step in LfO. It represents a general architecture for recognizing state-action pairs from RGB-D streams, which are then input to underlying methods for LfO such as inverse reinforcement learning~\cite{Arora19:Survey}. \sanet{} demonstrated recognition accuracies on two diverse LfO tasks that are significantly better than previous conventional techniques and a recent architecture that analyzes videos. This is expected to benefit the subsequent LfO. While minor changes in components may be beneficial, an ablation study revealed that the major architectural parts of the NN are indeed needed. A low resource utilization signature allows \sanet{} to be deployed on board robotic platforms. Though \sanet{} is shown to predict discretized state and action pairs, it could predict continuous ones by changing the classification network to a regression network.

\sanet{} brings another benefit to LfO. Recent techniques, such as maximum entropy deep inverse reinforcement learning~\cite{wulfmeier2015maximum}, utilize a NN. \sanet{} can be merged with the NN for inverse reinforcement learning, potentially producing the first end-to-end deep learning approach for LfO. As future work, \sanet{}'s performance could be improved by improving the LSTM~\cite{veeriah2015differential} or learning under weak supervision~\cite{richard2017weakly}.

\section*{Acknowledgements}

This research was partially supported by NSF grant \#IIS 1830421. We thank Kenneth Bogert for help with evaluation of the CMVision based baseline method. 

\bibliographystyle{abbrv}
\bibliography{shda19}

\end{document}